%% file: main.tex
\documentclass[10pt,twocolumn,letterpaper]{article}
\usepackage{pifont}
\usepackage[datasets]{wacv}
\input{preamble}
\definecolor{wacvblue}{rgb}{0.21,0.49,0.74}
\usepackage[pagebackref,breaklinks,colorlinks,allcolors=wacvblue]{hyperref}

\def\wacvPaperID{1541}
\def\confName{WACV}
\def\confYear{2027}

\title{REEL: Same Frames, Different Answers and Different Costs}

\author{Xin  Liang \quad Jiangpeng He    \\
Indiana University Bloomington\\
{\tt\small \{liangxin, jhe2\}@iu.edu}
}

\providecommand{\SelectorTableRows}{}
\input{selector_final_values}

\begin{document}
\maketitle
\input{sec/0_abstract}
\input{sec/1_intro}
\input{sec/2_related}
\input{sec/3_protocol}
\input{sec/4_experiments}

\input{sec/5_conclusion}
{
  \small
  \bibliographystyle{ieeenat_fullname}
  \bibliography{main}
}
\clearpage
\appendix
\input{sec/A_experimental_details}

\input{sec/B_additional_experiments} 
\end{document}

%% file: preamble.tex
\usepackage{booktabs}
\usepackage{multirow}
\usepackage{amsmath,amssymb}
\usepackage{enumitem}
\usepackage{xcolor}
\usepackage[most]{tcolorbox}
\usepackage{graphicx}
\usepackage{subcaption}
\usepackage{array}
\usepackage{makecell}
\usepackage{microtype}
\usepackage{xspace}
\usepackage{pifont}
\usepackage{colortbl}
\usepackage{placeins}
\usepackage{etoolbox}

\newlength{\tablebottomtrim}
\AtEndEnvironment{table}{\par\vspace{\tablebottomtrim}}
\AtEndEnvironment{table*}{\par\vspace{\tablebottomtrim}}

\newcommand{\method}{\textsc{REEL}\xspace}

\definecolor{evalblue}{RGB}{31,103,180}
\definecolor{evalred}{RGB}{190,58,52}
\definecolor{evalorange}{RGB}{220,126,34}
\definecolor{resultgood}{RGB}{226,242,231}
\definecolor{resultbad}{RGB}{250,228,228}
\definecolor{resultwarn}{RGB}{255,242,211}
\definecolor{resultinfo}{RGB}{228,239,249}
\newcommand{\goodcell}[1]{\cellcolor{resultgood}\textbf{#1}}

\definecolor{takeawayblue}{RGB}{78,103,124}
\definecolor{takeawaybg}{RGB}{247,249,252}
\newtcolorbox{takeawaybox}[1][Takeaway]{
  enhanced,
  width=\columnwidth,
  colback=takeawaybg,
  colframe=takeawayblue,
  boxrule=0.8pt,
  arc=3pt,
  left=6pt,
  right=6pt,
  top=11pt,
  bottom=6pt,
  before skip=9pt,
  after skip=9pt,
  attach boxed title to top left={xshift=5mm,yshift=-2.5mm},
  boxed title style={
    colback=takeawayblue,
    colframe=takeawayblue,
    arc=3pt,
    boxrule=0pt,
    left=6pt,
    right=6pt,
    top=2pt,
    bottom=2pt
  },
  fonttitle=\small\bfseries\color{white},
  fontupper=\small,
  title={#1}
}

%% file: selector_final_values.tex
\providecommand{\SelectorTableRows}{}

\renewcommand{\SelectorTableRows}{%
CG ($H=6$) & 32 & 38.5 & 46.0 & +7.6 & 63.7 & 65.4 & 65.1 & 199.7 \\
CG ($H=6$) & 64 & 41.9 & 48.1 & +6.2 & 62.6 & 65.4 & 101.5 & 236.2 \\
CG ($H=6$) & 128 & 48.1 & 54.7 & +6.7 & 61.9 & 65.4 & 214.7 & 349.8 \\
HD ($H=20$) & 32 & 32.4 & 30.9 & -1.5 & 196.8 & 222.9 & 168.0 & 938.5 \\
HD ($H=20$) & 64 & 31.3 & 30.2 & -1.1 & 196.2 & 222.9 & 281.7 & 1{,}049.3 \\
HD ($H=20$) & 128 & 30.9 & 30.8 & -0.1 & 195.4 & 222.9 & 565.3 & 1{,}345.2 \\
}

%% file: sec/0_abstract.tex
\begin{abstract}
 Video understanding keeps moving toward longer recordings, and the longer a recording is, the more questions  can be asked about it. But processing it again for every question does not scale, so many systems watch the video once, build a memory, and answer later questions from it. Yet what to remember is decided before anyone knows what will be asked, and whatever is left out cannot be recovered later. 
Moreover, whether the saving paid off is judged by token counts, which fall when the input gets shorter but not when the model reuses what it has already computed. We present\textbf{ \method}, an evaluation framework that compares three saved representations built from exactly the same observed frames, so that a change in answers can come only from what the system remembered. \textbf{Text} writes a short record of what happened, \textbf{Visual} keeps the encoder outputs, and \textbf{Prefix State} also keeps the language-model prefix. Each is compared with full reprocessing (\textbf{Frames}) by paired correctness changes, cumulative GPU time, and storage size. Across six vision-language models and two benchmarks, Text and Frames differ in correctness on 20\% to 29\% of questions, even when their mean accuracy is almost the same. By contrast,  Visual  and Prefix State stay much closer to Frames. With unlimited storage, Prefix State  is the cheapest in total GPU time among memories that keep almost all of the answers Frames gets right, for every model on both benchmarks. The trade-off is storage: Prefix State keeps up to about 21~GiB per video, whereas Text needs only kilobytes. Reusing computation and discarding information both look like efficiency, but only the second removes evidence that later questions may need, and neither a mean accuracy nor a token count tells them apart. \method makes that difference measurable for any system that answers many questions from one saved memory.
\end{abstract}

%% file: sec/1_intro.tex
\begin{figure}[t]
\centering
\begin{subfigure}[t]{0.485\linewidth}
  \centering
  \includegraphics[width=\linewidth]{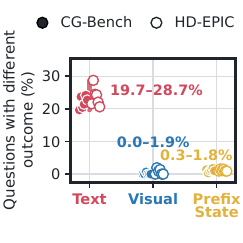}
  \subcaption{Correctness differences from Frames.}\label{fig:fm-answers}
\end{subfigure}\hfill
\begin{subfigure}[t]{0.485\linewidth}
  \centering
  \includegraphics[width=\linewidth]{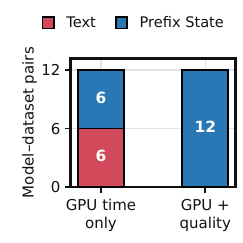}
  \subcaption{Representation selected by evaluation.}\label{fig:fm-selection}
\end{subfigure}
 
\begin{subfigure}[t]{0.485\linewidth}
  \centering
  \includegraphics[width=\linewidth]{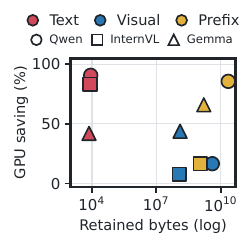}
  \subcaption{Storage versus saved GPU work.}\label{fig:fm-storage}
\end{subfigure}\hfill
\begin{subfigure}[t]{0.485\linewidth}
  \centering
  \includegraphics[width=\linewidth]{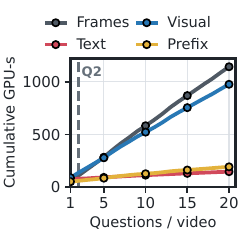}
  \subcaption{Cost over repeated questions.}\label{fig:fm-cost}
\end{subfigure}
\caption{\textbf{Representation changes answers, cost, and storage.}
(a) Correct/wrong changes across 12 model-dataset pairs, not mean accuracy loss.
(b) Lowest-GPU choice before and after answer-retention checks.
(c) Retained bytes versus GPU saving across three model families.
(d) Cumulative GPU cost on HD-EPIC.}
\label{fig:frontmatter-summary}
\end{figure}

\section{Introduction}
\label{sec:intro}

\begin{figure*}[t]
\centering
\includegraphics[width=\textwidth]{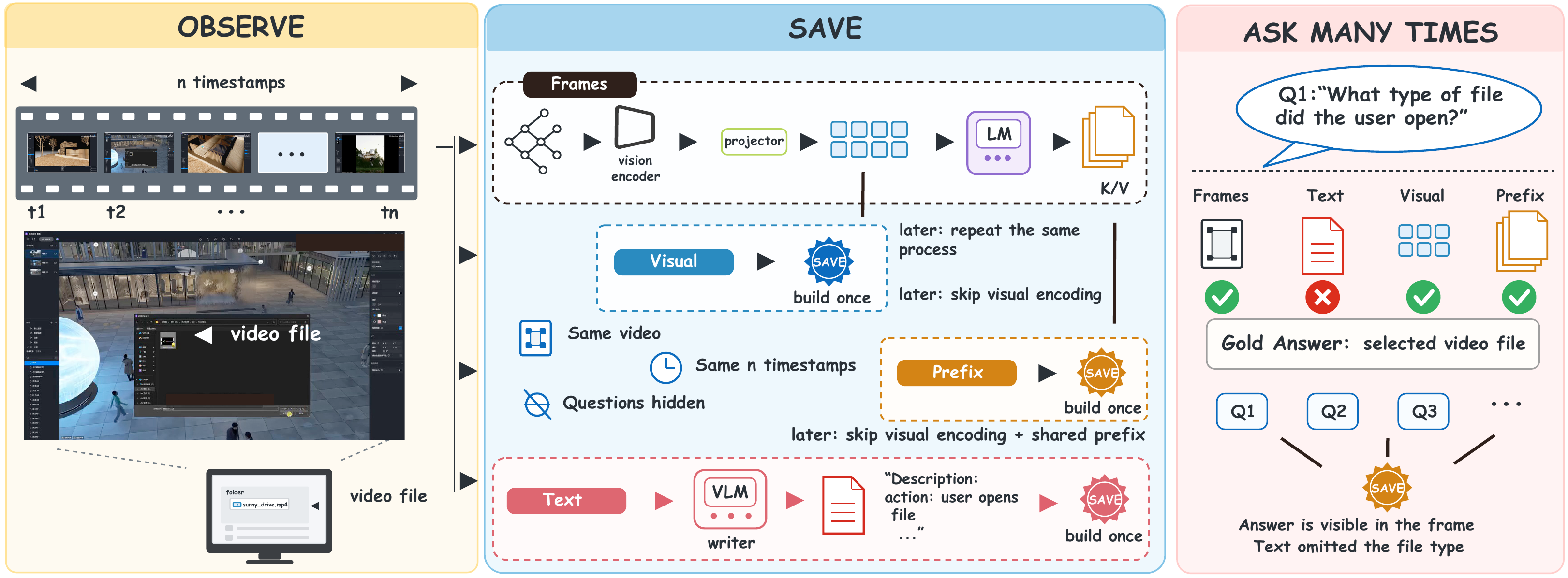}
\caption{\textbf{The missing video file.}
All four paths see the same timestamps. Frames repeats the full model;
Visual and Prefix State save intermediate model state; Text writes a
question-blind description. Here, the frame reveals a video file, but Text
records only ``user opens file,'' so its later answer differs.}
\label{fig:overview}
\vspace{-10pt}
\end{figure*}

A single long video can be queried with many different questions. For example, some long-video benchmarks  pair many questions with one video~\cite{song2024moviechat,chen2025cgbench,perrett2025hdepic}, and some
self-training pipelines keep generating new question-answer pairs from the same videos~\cite{chen2024bovila}.   A video may even be revisited months later with a question nobody had thought of at the time. In such a scenario, the video is often processed before the model knows what will be asked of it. Answering each question from scratch, however, means processing the video again,
and Figure~\ref{fig:fm-cost} shows how quickly this cost   grows as more questions arrive.   The obvious shortcut is therefore to watch the video once, store a memory of it, and answer every later question from that memory. However, the model gets only one chance to decide what the memory keeps, and every question that arrives afterwards must be answered from the memory alone. To understand what the memory should keep, we compare four ways to store a video. \textbf{Frames} keeps the image itself, \textbf{Text} writes
a description, \textbf{Visual} keeps the visual tokens produced by the vision encoder, and \textbf{Prefix State} keeps both these tokens and the key-value cache that the language model computes from them.

Figure~\ref{fig:overview} illustrates the four ways with a simple example. Here, a frame shows a user picking a file out of a folder. The Text writer has not seen any question yet, so it
records only ``user opens file.''  When a later question asks which kind of
file it was, the answer remains visible in the frame but is absent from the Text record. Frames, Visual, and Prefix State all answer this question correctly, but Text does not. What a system stores therefore determines not only how much computation it can avoid, but also which later questions it can still answer.

More generally, we distinguish two kinds of failure that can lead to a wrong answer: \ding{182} \textit{selector failure}, in which the system never sees the moment it needs; \ding{183} 
\textit{representation failure}, in which it sees that moment but does not keep the relevant detail.   The example in Figure~\ref{fig:overview} is a representation failure because the selector found the relevant frame but Text did not keep the file type.
Yet many long-video
evaluations give one final score to an end-to-end pipeline that both selects
frames and builds the saved representation, so the score cannot tell the two
failures apart~\cite{wang2023lifelongmemory,song2024moviechat,di2025rekv}.

Moreover, these evaluations often compare memories using two metrics, mean accuracy and token count, but both can be misleading.
Mean accuracy can stay the same while many individual answers change, because gains on some questions can cancel losses on others.
Token count stays the same whether the GPU processes the same input again or reuses saved results, so it cannot show how much computation is reused.
We therefore introduce \textbf{\method } (Repeated-query Evaluation of Efficiency in Long-video QA), an evaluation framework that gives all four ways the same selected frames, so that any change in an answer can come only from what was stored. Instead of mean accuracy and token count, \method compares each memory with Frames one question at a time and measures the GPU time to build the memory and to answer every later question. It also reports how many bytes the memory keeps.

Using \method, we find that Text stands apart from the other three.
Figure~\ref{fig:fm-answers} shows that Text and Frames differ in correctness on \textbf{20\% to 29\%} of questions across six models and two benchmarks, even when their mean accuracy is almost the same, whereas Visual and Prefix State stay much closer to Frames.
By GPU time alone, Text is the cheapest of the four in 6 of the 12 model-benchmark pairs, but on 10\% to 16\% of questions it is wrong where Frames is right.
Figure~\ref{fig:fm-selection} shows that with unlimited storage, Prefix State is the cheapest in all 12 pairs once a memory must keep almost all of the answers Frames gets right.
The price is storage, which Figure~\ref{fig:fm-storage} plots against the GPU time saved: across the three evaluated backbones, the median retained Prefix State size is at most about 21 GiB per video, whereas Text keeps only kilobytes.
Under a smaller storage budget, Visual and then Frames can take its place, as Figure~\ref{fig:storage-budget-selection} in Appendix~\ref{app:additional-results} shows.
We also find that frame selection has no single winner.
Table~\ref{tab:selector-sweep} shows that with Qwen3-VL-8B, AKS~\cite{tang2025aks}, which searches again for every question, beats MaxVol~\cite{li2026maxinfo}, which selects frames once per video, on CG-Bench but not on HD-EPIC, and that its repeated search raises total GPU time in every comparison.

This work makes three contributions.  First, \method  separates frame selection from
the representation saved for later questions, so that  selector failures and representation failures can be distinguished.  Second, we evaluate reuse
along three axes that can disagree: per-question answer changes, cumulative GPU time,
and storage per video. Third, across six vision-language models and two long-video question
benchmarks, we show that Text changes many more answers than Visual or Prefix State, even when mean accuracy is almost unchanged, and that the cheapest memory that keeps almost all answers depends on the storage budget.
Our code is available at \url{https://github.com/LiangXin1001/REEL-code}.

%% file: sec/2_related.tex
\section{Related Work}
\label{sec:related}

\subsection{Selecting Evidence and Storing It}

A long-video pipeline first chooses what to observe and then chooses what to
retain.  The first problem is \emph{evidence selection}; the second is
\emph{memory representation}.  Most systems solve both inside one pipeline,
where an end-to-end result cannot reveal whether the selector missed evidence
or the memory discarded it.

Frame selectors differ in when they see the question.  MaxInfo applies a
maximum-volume rule to visual embeddings without reading the question
\cite{li2026maxinfo}.  AKS balances prompt--frame relevance with temporal
coverage~\cite{tang2025aks}.  LongVU removes temporal redundancy and then uses
the text query to reduce frame features~\cite{shen2024longvu}.  Query-aware
selection can target the current question, but its question-dependent work
does not automatically serve the next one.
VideoAgent, VideoTree, LongVideo-R1, LensWalk, LongVT, and LongVideoAgent also
search or navigate after the question arrives
\cite{wang2024videoagent,wang2025videotree,qiu2026longvideor1,li2026lenswalk,
yang2026longvt,liu2026longvideoagent}.

Memory methods address what remains after observation.  Text systems store
activity logs or descriptions
\cite{wang2023lifelongmemory,islam2024videorecap,kahatapitiya2024langrepo}.
Visual systems retain embeddings or recurrent slots
\cite{song2024moviechat,he2024malmm,diko2025rewind,zhang2025flashvstream}.
Other systems preserve latent or KV state
\cite{yang2025streammem,chen2026flexmem,yeo2026worldmm}.  Text is a lossy
semantic record; visual and KV state preserve intermediate model results.
They expose different information, skip different work, and occupy different
memory tiers, so input length alone cannot compare them.

ReKV encodes a video before questions arrive, stores video KV on RAM or disk,
and retrieves relevant state for each question~\cite{di2025rekv}.  Ong and
Jang construct one multimodal prefix KV cache from uniformly sampled episode
frames and reuse it across all questions in that episode~\cite{ong2025episode}.
Dive compresses visual KV once before future questions arrive and reports
cumulative latency as this one-time cost is amortized~\cite{hong2026dive}.
NKR instead distills each video once into reusable lightweight network
weights~\cite{guan2026nkr}.  These studies optimize particular reusable
representations.  REEL instead holds the observations and backbone fixed and
compares multiple save points through paired answer transitions, cumulative
GPU work, and retained bytes.

KFS-Bench studies whether selected frames cover the needed scenes.
MemoryCard tests which fields survive in event cards, and VideoSEAL separates
access to the right time from semantic support for an answer
\cite{li2026kfsbench,yang2026memorycard,qiu2026videoseal}.  Following this
distinction, we give every memory form the same selected timestamps.  Any
answer difference then comes from what the memory keeps, rather than from which
frames it saw.

\subsection{Evaluating Reuse across Multiple Questions}

Long-video benchmarks increasingly contain several questions for one video.
NExT-GQA and CG-Bench test temporal grounding
\cite{xiao2024nextgqa,chen2025cgbench}; HERBench and CaST-Bench emphasize
distributed or causal evidence~\cite{ami2026herbench,zhang2026castbench}; and
HD-EPIC covers dense egocentric questions~\cite{perrett2025hdepic}.
Video-MME-v2 also groups related questions~\cite{fu2026videommev2}.  Their
standard protocols score each answer, but do not charge shared video
processing across an ordered sequence of questions.

Caption-based and cache-based studies cover different parts of this problem.
CaptionQA, CapQuiz, and CAPability test which visual facts survive in language
\cite{yang2026captionqa,wang2026capquiz,liu2025capability}.  CapMem directly
treats captions as reusable episodic memory, includes a matched-frame
CaptionQA--VideoQA control, and amortizes caption construction over multiple
questions~\cite{liang2026capmem}.  Unlike these studies, REEL jointly compares
Text, retained visual features, and reusable prefix state within one matched
pipeline and reports directional paired answer transitions, cumulative GPU
work, and retained bytes.  VLCache, Kamera, and MuKV measure the latency or
size of reused model state~\cite{qin2025vlcache,kamera2026,xiao2026mukv}.

Our evaluation connects these two views.  We compare Text, Visual, and Prefix
State inside the same model pipeline, using the same timestamps, questions,
answer format, and hardware.  We then ask which answers change, how much GPU
work each path avoids over repeated questions, and how much state it keeps.

%% file: sec/3_protocol.tex
\section{REEL Evaluation Protocol}
\label{sec:protocol}

As Figure~\ref{fig:overview} illustrates, all four representations start from
the same observations but retain different information and completed
computation.  We now define these representations and how they are evaluated.

\subsection{What Is Selected and Saved}

Each evaluation unit is one video followed by several questions.  A
question-blind selector chooses its timestamps before seeing any question and
reuses that list for the whole video.  A question-aware selector may read the
current question, but must run again when the question changes.

RQ1--RQ2 use 64 uniformly spaced timestamps so that every representation sees
exactly the same observations.  Uniform sampling is a shared input, not a
claim about the best selector.  We repeat the representation comparison with
\textbf{MaxInfo-derived Chunked MaxVol}, a content-aware, question-blind
selector based on MaxInfo's chunked maximum-volume idea~\cite{li2026maxinfo}.
Our fixed-budget version is not an exact reproduction of MaxInfo, which may
return a variable number of frames.  RQ3 compares Chunked MaxVol with
question-aware AKS~\cite{tang2025aks} at 32, 64, and 128 frames.  MaxVol runs
once per video; AKS runs once per question.  We include the complete cost each
time either selector runs.  Appendix~\ref{app:details} gives both algorithms
and their settings.

Given the same timestamps, we compare four answer paths.  \textbf{Frames}
keeps no model state and processes the selected frames again for every
question.  \textbf{Text} uses the evaluated backbone in a one-time,
question-blind construction stage to write a bounded record, then uses the
same backbone to answer from that record.
An unwritten detail is no longer explicitly available.  \textbf{Visual} retains
the encoder and connector
outputs, avoiding repeated visual encoding.  \textbf{Prefix State} also keeps
the KV pages created during prefix prefill~\cite{kwon2023vllm}, avoiding both
visual encoding and most repeated prefill.  These handoffs preserve different information and
different completed computation; they are not merely different file formats.
We use these four names throughout; ``KV'' refers only to the cache mechanism
inside Prefix State.

We also test Text+Visual and Text+Prefix State in one answer call.  The Text
record is the same one used by Text alone, and its writing cost is counted
once.  Prefix State already contains the retained Visual state, so adding
Visual to it supplies no additional evidence.  Supplying all three is
therefore the same condition as Text+Prefix State.  Execution traces verify
that Visual does not rerun the encoder, and server counters verify prefix
reuse.  We test these combinations only in the three-model CG-Bench MaxVol-64
diagnostic, not in the 12-model-dataset pair representation matrix.

\subsection{What We Measure}

For a question with valid outputs from both paths, let $c^0$ be correctness
with Frames and $c^r$ correctness with representation $r$.  We measure changes
in both directions:
\begin{equation}
\begin{aligned}
\mathrm{Loss}_r &= \Pr(c^0=1,c^r=0), &
\mathrm{Gain}_r &= \Pr(c^0=0,c^r=1),\\[-1pt]
\Delta\mathrm{Acc}_r &= \mathrm{Gain}_r-\mathrm{Loss}_r, &
\mathrm{Flip}_r &= \mathrm{Gain}_r+\mathrm{Loss}_r.
\end{aligned}
\label{eq:answer-transitions}
\end{equation}
Here, Loss is the share of jointly valid questions on which Frames is correct
and representation $r$ is wrong; Gain is the share that changes in the
opposite direction.  Their difference is the usual mean accuracy change; their
sum is the share whose correct/wrong status changes.  Mean accuracy can
therefore remain stable while many individual answers change.
Appendix~\ref{app:details} defines valid outputs and the paired bootstrap. Conditional on the jointly valid set, our Flip is the same paired
correct$\leftrightarrow$incorrect event used by Dutta et al. to evaluate
compressed LLMs~\cite{dutta2024accuracy}. We report its two directions
separately because Loss measures answers retained by Frames but discarded by
representation $r$, whereas Gain measures the reverse transition.

The cumulative allocated GPU time after $m$ questions is
\begin{equation}
G_v^r(m)=B_v^r+\sum_{i=1}^{m}P_{v,i}^r,
\label{eq:cumulative-metric}
\end{equation}
where $B_v^r$ is work performed once for video $v$, including question-blind
selection and representation construction.  $P_{v,i}^r$ is the cost of the
$i$th answer.  For AKS, selection is part of every $P_{v,i}^r$ because it runs
again for every question.  Shared work is counted once; repeated work is
counted each time it runs.

Dividing the cumulative cost by $m$ reduces the one-time term to $B_v^r/m$.
More questions can therefore make reuse cheaper on average, but they cannot
restore information that was not saved.  One GPU-second is the number of
allocated GPUs multiplied by active execution time.  We compare GPU time only
under the same hardware and inference settings, and report wall time, tokens,
and monetary cost separately.

Storage is also a separate measurement.  We count the unique state retained
after construction, not temporary working memory.  Text uses its serialized
UTF-8 payload; Visual uses the backing allocations of its retained encoder
outputs; Prefix State includes the Visual backing and the logical KV pages of
the reusable prefix.  Aliases of the same allocation count once, and host and
GPU bytes are reported separately.  Candidate features and scores discarded
after selection contribute runtime and peak memory, but not retained size.
Appendix~\ref{app:details} gives the exact accounting rules.

\subsection{When Reuse Is Useful}

We do not call a representation useful merely because it crosses the Frames
cost curve.  At a measured question horizon, it must pass three checks:
\begin{enumerate}[leftmargin=18pt,itemsep=1pt,topsep=2pt]
\item its cumulative GPU time is no greater than Frames;
\item the lower 95\% bootstrap bound on mean accuracy change is at least
  $-3$ percentage points; and
\item no more than 3\% of jointly valid questions change from Frames-correct
  to representation-wrong.
\end{enumerate}
We report the first measured horizon at which all three checks pass and remain
passed at every later measured horizon.  The final requirement prevents an
early crossing from hiding a later question that needs information the
representation discarded.  The third check describes retention on the
evaluated questions; it is not a population confidence bound.  The appendix
gives the formal definition, crossing uncertainty, and sensitivity to both
quality limits.

%% file: sec/4_experiments.tex
\section{Experiments}
\label{sec:experiments}

\subsection{Experimental Setup}
\label{sec:setup}

CG-Bench MQ-50~\cite{chen2025cgbench} has 50 videos, each followed by six
ordered questions.  HD-EPIC-51~\cite{perrett2025hdepic} has 51 videos with 20
to 86 questions each.  Every path attempts
every available question.  HD-EPIC cumulative curves stop after question 20 so
that every video contributes at each point.

We test Qwen3-VL-8B and Qwen3-VL-4B~\cite{bai2025qwen3vl},
InternVL3-8B~\cite{zhu2025internvl3} and
InternVL3.5-8B~\cite{wang2025internvl35},
Gemma-3-12B-IT~\cite{gemmateam2025gemma3}, and
MiniCPM-V-4.5-8B~\cite{yu2026minicpmv45}.  The six backbones and two datasets give
12 model--dataset pairs, each run on one NVIDIA L40S GPU.  Appendix
Table~\ref{tab:model-runtime-config} lists the checkpoints.  We measure stored
size for Qwen3-VL-8B, InternVL3-8B, and Gemma-3-12B-IT; Table~\ref{tab:cost-profile}
reports their costs, and Appendix~\ref{sec:new-model-rq2} reports the other three.

For RQ1 and RQ2, every representation starts from the same 64 uniformly
sampled timestamps, chosen before any question is shown.
\textsc{Text}, \textsc{Visual}, and \textsc{Prefix State} build their reusable
states from those timestamps without seeing the questions.  After each
question arrives, \textsc{Frames} processes the same sampled frames from
scratch.  Each comparison uses the same questions, answer
format, and hardware.  The prompt changes only where the representation
requires a different input.  Appendix~\ref{app:details} gives the full protocol.

We compare answers only when both outputs contain exactly one valid option
label; every attempted call remains in the cost.
Appendix~\ref{sec:answer-format-exclusions} defines the excluded outputs,
reports their frequency, and checks the alternative of counting them as wrong.

\subsection{RQ1: Which Individual Answers Change?}
\label{sec:rq1}

\begin{figure}[t]
\centering
\begin{subfigure}[t]{\linewidth}
  \centering
  \includegraphics[width=\linewidth]{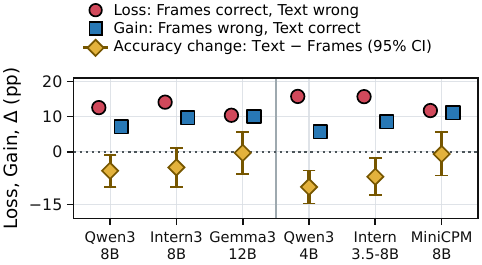}
  \subcaption{CG MQ-50: 300 questions.}
\end{subfigure}
 \vspace{10pt}
\begin{subfigure}[t]{\linewidth}
  \centering
  \includegraphics[width=\linewidth]{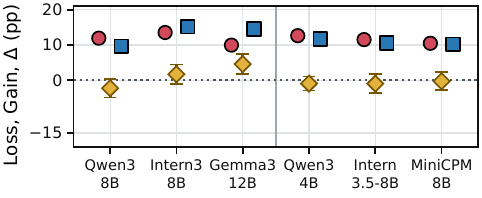}
  \subcaption{HD-EPIC-51: 1,871 questions.}
\end{subfigure}
\caption{\textbf{Many answers change even when mean accuracy changes little.}
Each model contributes one red circle (Loss) and one blue square (Gain) from
Eq.~\ref{eq:answer-transitions}; their sum is Flip. Yellow diamonds show
$\Delta\mathrm{Acc}$ (Text minus Frames) with 95\% video-bootstrap intervals. Points are model-level summaries, not individual
questions.}
\label{fig:rq1-transitions}
\vspace{-10pt}
\end{figure}

Mean accuracy counts correct answers, but it does not tell us whether two
conditions answer the same questions correctly.  We pair every Text prediction
with Frames and report Loss, Gain, and Flip from
Eq.~\ref{eq:answer-transitions}.

\paragraph{A stable mean can hide a different answer set.}
InternVL3-8B on HD-EPIC makes the cancellation easy to see.  Text and Frames
differ in mean accuracy by less than two points, yet Text's Flip is 28.7\%.
Its Loss is 13.5\% of jointly valid pairs and its Gain is 15.2\%, so the two
directions nearly cancel in the mean.  The same pattern appears across models
in Figure~\ref{fig:rq1-transitions}.  By
contrast, the Flip of Visual and Prefix State is at most 2\% in all 12
model--dataset pairs.  Because all paths see the same timestamps, these
differences arise after frame selection.
Across these pairs, the smallest lower end of the 95\% video-bootstrap
interval for Text's Flip is 14.8\%, as Appendix~\ref{sec:text-flip-uncertainty}
reports.

\paragraph{Writing priorities change Text outcomes.}
Balanced24 asks the writer to record entities, actions or states, and visible
text in at most 24 words per frame.  Detail24 instead prioritizes contents,
counts, colors, spatial relations, labels, and exact visible text.  Everything
else is the same, including the frames, schema, word limit, temperature, and
output ceiling.  Changing only the priority instruction alters realized length
by just 0.4--2.8\%, yet the Flip between the two writers is 6.8--9.8\%.  The
writer's priorities therefore affect which details the Text record preserves.
Appendix Table~\ref{tab:prompt-ablation-full} gives the paired changes,
bootstrap intervals, and valid-output rates.  Adding a whole-video summary of at most 128 words to Balanced24 does not
close the gap with Frames: Flip remains 18.6--29.6\% and Loss remains
9.7--17.0\%. Further details are provided in Appendix
Section~\ref{sec:global-summary-method}.

\begin{figure}[t]
\centering
\includegraphics[width=\columnwidth]{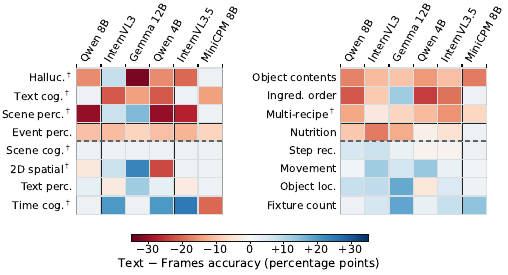}
\caption{\textbf{Text-minus-Frames accuracy by question type.}
Left: CG-Bench; right: HD-EPIC. Each shows the four lowest and four highest
model-averaged types; positive values favor Text, and $^\dagger$ marks types
with fewer than 30 questions.}
\label{fig:type-heatmap-main}
\vspace{-10pt}
\end{figure}

\paragraph{The changes depend on what the question asks.}
Figure~\ref{fig:type-heatmap-main} shows that the type-level pattern differs
between the two benchmarks.  On HD-EPIC, Text consistently hurts
questions about object contents and helps questions about movement across all
six backbones.  The full permutation test finds this dependence on question
type for HD-EPIC but not for CG-Bench, as Appendix~\ref{sec:question-types}
reports.  The test shows an association over question types, not a causal
account of individual predictions.

Question frequency can also change the aggregate conclusion.  Gemma appears
better with Text under HD-EPIC's observed mix (+4.6 points), but not when each
question type receives equal weight ($-0.1$ points).  A benchmark's type
distribution can therefore make a representation effect appear or disappear.

\paragraph{Both paths improve when they observe better evidence.}
Moving the sampling positions into CG-Bench's annotated clue windows improves
both Frames and Text.  As Appendix Table~\ref{tab:evidence-window-full} shows,
it does not consistently narrow the gap between them; all three interaction
intervals include zero.  Selection and preservation should therefore be
evaluated separately.  This diagnostic uses the current question's annotation, so
it is not a reusable selector and is not included in RQ2.

\begin{takeawaybox}[RQ1 answer]
\textbf{With the sampled frames held fixed, Text and Frames differ in
correctness on 19.7--28.7\% of questions across the 12 model--dataset
pairs, while the corresponding rates for Visual and Prefix State are at
most 2\%.}
Loss and Gain identify which Frames-correct answers are lost and which
Frames-wrong answers are recovered.
\end{takeawaybox}

\subsection{RQ2: When Does Reuse Become Cheaper and Pass the Quality Checks?}
\label{sec:rq2}

We now ask when reuse becomes cheaper without discarding too many answers.
Section~\ref{sec:protocol} defines three checks against Frames: cumulative GPU
time, $\Delta\mathrm{Acc}$, and Loss.  Text often becomes cheaper and can look
acceptable in the mean because its Gain offsets its Loss.  At
the default limits, Prefix State is the only representation that remains
eligible in every one of the 12 model--dataset pairs.
Appendix~\ref{sec:quality-threshold-sensitivity} reports the
per-representation counts and tests alternative quality limits.

\begin{table}[t]
\centering
\caption{\textbf{Lowest-cost eligible representation as questions arrive.}
``+ Quality'' applies both quality checks. F/T/V/P denote Frames, Text, Visual,
and Prefix State; subscripts mark the first question after a change.}
\label{tab:selection-paths}
\scriptsize
\setlength{\tabcolsep}{2.0pt}
\resizebox{\columnwidth}{!}{\begin{tabular}{lcccc}
\toprule
& \multicolumn{2}{c}{CG, $H=6$} & \multicolumn{2}{c}{HD, $H=20$} \\
\cmidrule(lr){2-3}\cmidrule(l){4-5}
Model & GPU only & + Quality & GPU only & + Quality \\
\midrule
Qwen3-VL-8B & F$\to$P$_{1}$ & F$\to$P$_{1}$ & F$\to$P$_{2}$$\to$T$_{7}$ & F$\to$P$_{2}$ \\
InternVL3-8B & F$\to$T$_{3}$ & F$\to$P$_{5}$ & F$\to$T$_{3}$ & F$\to$P$_{5}$ \\
Gemma-3-12B-IT & F$\to$P$_{1}$ & F$\to$P$_{2}$ & F$\to$P$_{2}$ & F$\to$P$_{2}$ \\
Qwen3-VL-4B & F$\to$P$_{2}$ & F$\to$V$_{4}$$\to$P$_{6}$ & F$\to$P$_{2}$$\to$T$_{4}$ & F$\to$P$_{2}$ \\
InternVL3.5-8B & F$\to$T$_{4}$ & F$\to$P$_{6}$ & F$\to$T$_{3}$ & F$\to$P$_{5}$ \\
MiniCPM-V-4.5-8B & F$\to$P$_{2}$ & F$\to$P$_{2}$ & F$\to$P$_{2}$ & F$\to$P$_{2}$ \\
\bottomrule
\end{tabular}}
\end{table}

Table~\ref{tab:selection-paths} shows that breaking even early does not
identify the representation that remains best.
With GPU time alone, the endpoint choice alternates between Text and Prefix
State.  Once answer retention is checked and storage is unconstrained, Prefix
State is the final choice in all 12 pairs.  Visual appears briefly in one
CG-Bench run because it meets the quality criteria before Prefix State does.
The checks therefore change the chosen representation, not only its crossing
point.  Two CG choices are close: Prefix State saves 3.0\% and 0.9\% of Frames'
GPU time, and both bootstrap intervals include zero.

\begin{table}[t]
\centering
\caption{\textbf{GPU time by model and question horizon.}
Build is one-time construction cost, Query is mean per-question answer cost,
and Total is cumulative GPU-seconds per video. The HD columns additionally
report the all-question total and its saving relative to Frames. Within each model--dataset block, green shading indicates the representation
with the lowest cumulative GPU cost.}
\label{tab:cost-profile}
\scriptsize
\setlength{\tabcolsep}{1.6pt}
\renewcommand{\arraystretch}{0.96}
\resizebox{\columnwidth}{!}{%
\begin{tabular}{lrrr rrrrr}
\toprule
& \multicolumn{3}{c}{CG} & \multicolumn{4}{c}{HD} & \\
\cmidrule(lr){2-4}\cmidrule(lr){5-8}
Rep. & Build & Query & Total$_6$ & Build & Query$_{1:20}$ & Total$_{20}$ & Total$_{\rm All}$ & Saving$_{\rm All}$ \\
\midrule
\multicolumn{9}{l}{\tiny\itshape Qwen3-VL-8B} \\[0.8pt]
Frames & 0.0 & 13.2 & 79.5 & 0.0 & 57.2 & 1,143.1 & 2,084.5 & 0.0\% \\
Text & 60.4 & 3.1 & 79.3 & \goodcell{74.0} & \goodcell{3.6} & \goodcell{146.5} & \goodcell{193.4} & 90.7\% \\
Visual & 9.5 & 10.0 & 69.5 & 47.9 & 46.4 & 976.3 & 1,737.6 & 16.6\% \\
Prefix & \goodcell{9.5} & \goodcell{3.9} & \goodcell{32.8} & 46.9 & 7.3 & 192.5 & 298.5 & 85.7\% \\
\midrule
\multicolumn{9}{l}{\tiny\itshape InternVL3-8B} \\[0.8pt]
Frames & 0.0 & 18.3 & 110.1 & 0.0 & 18.2 & 363.7 & 667.9 & 0.0\% \\
Text & \goodcell{46.2} & \goodcell{1.8} & \goodcell{57.0} & \goodcell{41.7} & \goodcell{1.9} & \goodcell{79.8} & \goodcell{111.4} & 83.3\% \\
Visual & 13.6 & 17.0 & 115.4 & 14.3 & 16.7 & 347.5 & 616.6 & 7.7\% \\
Prefix & 13.6 & 15.5 & 106.8 & 14.4 & 14.9 & 313.1 & 556.6 & 16.7\% \\
\midrule
\multicolumn{9}{l}{\tiny\itshape Gemma-3-12B-IT} \\[0.8pt]
Frames & 0.0 & 10.5 & 62.8 & 0.0 & 10.5 & 210.0 & 378.6 & 0.0\% \\
Text & 73.1 & 3.8 & 95.9 & 67.3 & 4.5 & 156.5 & 219.4 & 42.1\% \\
Visual & 7.3 & 5.6 & 40.7 & 8.2 & 5.8 & 123.2 & 212.8 & 43.8\% \\
Prefix & \goodcell{7.3} & \goodcell{3.3} & \goodcell{27.2} & \goodcell{8.3} & \goodcell{3.5} & \goodcell{77.6} & \goodcell{128.8} & 66.0\% \\
\bottomrule
\end{tabular}}
\end{table}

Table~\ref{tab:cost-profile} reports the main cost values.  The full cumulative
curves and bootstrap intervals are shown in Appendix
Figure~\ref{fig:cumulative-cost-full}; Figure~\ref{fig:representation-size-full}
adds retained size and $\Delta\mathrm{Acc}$.  Text usually has the lowest answer
cost, but its one-time construction can outweigh that advantage when a
video receives only a few questions.  Prefix State costs about as much to
build as Visual yet avoids more work on each later question.  The cheapest
path depends on the model, construction cost, and number of questions.

\begin{figure}[t]
\centering
\begin{subfigure}[t]{.49\linewidth}
  \centering
  \includegraphics[width=\linewidth]{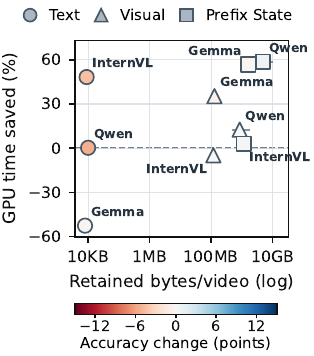}
  \subcaption{CG MQ-50, six questions.}
\end{subfigure}\hfill
\begin{subfigure}[t]{.49\linewidth}
  \centering
  \includegraphics[width=\linewidth]{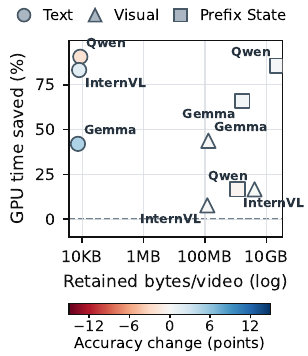}
  \subcaption{HD-EPIC-51, all questions.}
\end{subfigure}
\caption{\textbf{Storage, GPU saving, and accuracy.}
Position gives bytes and GPU saving; color gives $\Delta\mathrm{Acc}$ against
Frames.  Prefix State includes Visual storage and KV pages. Qwen, InternVL, and Gemma denote Qwen3-VL-8B, InternVL3-8B,
and Gemma-3-12B-IT, respectively.}
\label{fig:representation-size-full}
\vspace{-10pt}
\end{figure}

\paragraph{A high prefix-hit rate does not determine GPU savings.}
Visual and Prefix State use nearly the same logical prompt, but only Prefix
State reuses the language-model prefix.  A high hit rate records how much of
that prefix is reused, not how costly it was to compute; end-to-end savings
therefore differ widely.  Visual reports no language-model cache hits because
it reuses encoder outputs rather than language-model state.

\paragraph{Prefix State saves more work for a similar construction cost.}
Visual and Prefix State have similar measured construction times; both
construction paths first compute visual state.  Visual reuses encoder outputs,
while Prefix State also reuses the language-model prefix.  Prefix State is
therefore cheaper per question and, without a storage limit, is the final
answer-preserving choice throughout our study.  For the three profiled
backbones, Visual retains less state.  Across all six, it is cheaper than Frames
while passing both quality checks in 10 of the 12 model--dataset pairs; the
other two fail only the cost condition.

\paragraph{Construction cost is amortized over repeated questions.}
Dividing Eq.~\ref{eq:cumulative-metric} by the number of questions reduces the
one-time construction term to $B/m$.  It does not change the answers stored in
the representation or reduce its Loss.  A Text path can therefore cross the
Frames cost curve without passing the quality checks.  Prefix State avoids
this gap by reusing computation without first rewriting the visual observation
as text.

\paragraph{Storage budgets change the selected representation.}
For each model--dataset--representation setting, we compare the median
retained size per video with the storage budget. A representation is
considered feasible when this median does not exceed the budget. We then
apply the GPU-time and quality checks, using Frames as the fallback.
Original-video storage is excluded from this analysis, so Frames adds no
retained representation state.

\paragraph{Retained size and GPU time rank the paths differently.}
For the three profiled backbones, the median saved representation ranges from
kilobytes for Text to 1.06--20.97 GiB for a full Prefix State, as
Figure~\ref{fig:representation-size-full} shows.  The smallest representation
is not the cheapest one that preserves answers.  Saved size, GPU time, and
answer quality measure different properties.

\paragraph{Token pricing favors input shortening unless it recognizes internal
reuse.}
Token counts and GPU time describe different parts of the same run.  On
HD-EPIC with Qwen3-VL-8B, Prefix State reports slightly more prompt tokens than
Frames yet uses much less GPU time.  It saves work by reusing prefix prefill
without shortening the logical input.  Text shortens the input itself, so a
token ledger records its saving directly.

Applying a published token tariff can rank the same runs differently because
the bill may not credit reused visual encoding or local prefix reuse.  Appendix
Table~\ref{tab:glm-tariff-crossing} applies the published GLM
prices~\cite{zhipu2026pricing,zhipu2026cache} to our recorded token counts.  The
calculation re-prices our local ledgers; it is not a provider bill.

We therefore report these measurements separately: paired outcomes show which
answers change, GPU time records executed work, and token counts describe the
logical prompt.

\begin{takeawaybox}[RQ2 answer]
\textbf{With unconstrained storage, Prefix State is the lowest-GPU
representation that passes both quality checks in all 12 model--dataset
pairs; under the GPU-only criterion, Text is selected in six.}
Comparisons of video memories should report Loss and retained bytes alongside
cumulative GPU time.
\end{takeawaybox}

\subsection{RQ3: What Does Question-Aware Selection Gain and Cost?}
\label{sec:rq3}

RQ1--RQ2 compare representations on the same frames.  RQ3 changes the frame
selector instead.  The full $K\in\{32,64,128\}$ sweep uses Qwen3-VL-8B with
the Frames answer path on CG-Bench and HD-EPIC, so a saved representation
cannot explain a difference between selectors.  At each budget, the answer
model and inference settings are the same.  To test transfer beyond Qwen3-VL-8B,
Appendix~\ref{sec:selector-transfer-k64} repeats the matched $K=64$ comparison
across all six backbones.

Chunked MaxVol scans the 1-Hz stream with a pretrained CLIP model once per
video, before any questions arrive.  Its selected timestamps are reused for
every question about that video.  AKS rescans the stream and uses a pretrained
BLIP~\cite{li2022blip} model to score frames after each question arrives.  We include video
decoding, scoring, and recursive selection every time AKS runs.  Although the
sweep code shares some work across frame budgets, the reported cost treats
each budget as an independent run.

As a simple reference, we also compare AKS with uniform sampling on CG-Bench.
AKS is better at every frame budget when invalid answers count as wrong, as
Appendix Table~\ref{tab:uniform-aks-reference} shows.  This verifies that its
gain is not tied to one particular frame budget.  The main comparison below is
harder: MaxVol is also content-aware, but its video scan can be reused.

The MaxVol--AKS comparison instead uses questions with valid answers from both
selectors.  Paired-valid coverage is at least 95.0\%, and every attempted call
remains in the cost.

Table~\ref{tab:selector-sweep} reports accepted runs only.  Selector time
includes the full video scan, and total GPU time includes both selection and
answering over the first 6 CG or 20 HD questions of each video.  MaxVol pays
for selection once per video; AKS pays again for every question.

\begin{table}[t]
\centering
\caption{\textbf{MaxVol and AKS at the same frame budgets with Qwen3-VL-8B.}
MaxVol is question-blind and AKS is question-aware; costs include selection
and answering.}
\vspace{-8pt}
\label{tab:selector-sweep}
\tiny
\setlength{\tabcolsep}{1.2pt}
\renewcommand{\arraystretch}{0.94}
\resizebox{\columnwidth}{!}{%
\begin{tabular}{llrrr@{\hspace{3pt}}rr@{\hspace{3pt}}rr}
\toprule
& & \multicolumn{3}{c}{Acc. (\%)} &
\multicolumn{2}{c}{Select (s)} & \multicolumn{2}{c}{Total GPU-s} \\
\cmidrule(lr){3-5}\cmidrule(lr){6-7}\cmidrule(l){8-9}
Data & $K$ & MaxVol & AKS & $\Delta$ & MaxVol/video & AKS/question & MaxVol & AKS \\
\midrule
\SelectorTableRows
\bottomrule
\end{tabular}}

 \vspace{-10pt}
\end{table}

\begin{takeawaybox}[RQ3 answer]
\textbf{With Qwen3-VL-8B, AKS is 6.2--7.6 points more accurate than MaxVol
on CG-Bench and 0.1--1.5 points less accurate on HD-EPIC.}
Because AKS searches once per question, its total GPU time is
$1.63$--$5.59\times$ that of MaxVol; selector comparisons should include the
cost of every search.
\end{takeawaybox}

\subsection{Robustness and Scope}
\label{sec:robustness}

\paragraph{What remains stable.}
The main result is qualitative: changing the saved representation can change
which questions are answered correctly even when the average accuracy moves
very little.  We see this pattern across the tested model families and on both
main datasets.  It also remains after changing the Text writer's priorities,
adding a video-level summary, changing the order in which questions arrive,
and extending the question horizon on HD-EPIC.  Repeating the same path with
the same prompt produces much smaller answer changes than switching from
Frames to Text (Appendix~\ref{sec:run-to-run}).  Our paired correctness analysis
is limited to multiple-choice questions with a definite correct answer. We do
not assume that the same metric transfers unchanged to open-ended multimodal generation~\cite{liang2026slidegen}.

The finding is not tied to one exact frame budget.  Using fewer observed frames
changes some answers, as expected, but it does not change the main comparison:
Text departs more from Frames, whereas Visual and Prefix State remain much
closer.  Separately, we test the selector comparison at several frame budgets
and repeat a matched-budget comparison across backbones.  Appendix
Table~\ref{tab:uniform32} and Section~\ref{sec:selector-transfer-k64} report the
detailed results.

\paragraph{The pattern is not specific to uniform sampling.}
We also repeat the representation comparison after Chunked MaxVol selects the
frames.  Across the tested backbones, Text again changes substantially more
answers than either retained-state path.  Adding Text to Visual or Prefix State
does not provide a consistent improvement.  This check shows that the main
observation is not simply an artifact of the uniform sampler: after the same
frames have been selected, the form in which they are retained still matters.
Appendix Section~\ref{sec:maxvol-representation-check} gives the paired results
and applies the same RQ2 retention rule.

\paragraph{A third benchmark changes the question mix.}
HER-MQ10, drawn from HERBench Lite-v2~\cite{ami2026herbench}, emphasizes action
sequences, false-action memory, and temporal shot ordering.  We use it as a
separate transfer check rather than folding it into the main summary.  The
result tells the same practical story.  Text is compact and inexpensive, but
it discards too many answers that Frames gets right.  Visual and Prefix State
stay closer to Frames, and Prefix State is the lowest-cost path that passes the
retention rule.  Appendix Table~\ref{tab:herbench-full} gives the complete
paired results.

%% file: sec/5_conclusion.tex
\section{Conclusion}

A frame selector decides what a video system sees. A representation decides
what remains available when the next question arrives.  \method separates
these decisions so that a missing observation is not confused with a lost
detail. The representation study reveals a trade-off that mean accuracy alone misses.
Text is compact and often cheap to query, but it can exchange many correct
answers for different ones.  Visual and Prefix State behave more like Frames.
Prefix State also avoids language-model prefill, so it can save GPU work
without shortening the input.  Repeated questions repay a one-time
construction cost; they cannot restore a detail that was never written. Evaluations of repeated video QA should therefore report which frames were selected, what was saved, which individual answers changed, how much GPU work
actually ran, and how much state remained.  No one of these measurements can
stand in for the others.

%% file: sec/A_experimental_details.tex
 
\setcounter{dbltopnumber}{3}
\setcounter{topnumber}{3}
\renewcommand{\dbltopfraction}{0.9}
\renewcommand{\topfraction}{0.9}
\renewcommand{\textfraction}{0.05}
\renewcommand{\floatpagefraction}{0.85}
\renewcommand{\dblfloatpagefraction}{0.85}

\section{Detailed Experimental Protocol}
\label{app:details}

\subsection{Evaluation Cohorts}

We fixed every cohort before running any model, and model outputs play no role
in its construction.  CG MQ-50 is drawn from the locally playable CG-Bench
videos that also have subtitle files.  A deterministic round-robin with seed
\texttt{20260722} balances duration and the majority question type while
limiting questions per video.  Subtitle availability affects cohort membership,
but no path receives subtitle content.  HD-EPIC-51 uses only official questions
whose input is one complete video.  We select videos round-robin across
participants, keep every eligible question of a selected video, and order the
questions by a second round-robin over official question prototypes.  The two
main cohorts contain 300 of the 12,129 released CG-Bench questions and 1,871 of
the 2,730 released HD-EPIC questions with a single full-video input.  HER-MQ10
takes the first 50 lexicographically sorted video IDs of HERBench
Lite-v2~\cite{ami2026herbench} with at least ten valid timestamped questions and
keeps the first ten valid questions of each video.  We evaluate it with
Qwen3-VL-8B and InternVL3-8B and report it separately from the six-model
summary.

\subsection{Text Records and Answer Prompt}

For every model--dataset pair, the same checkpoint writes the Text record and
answers the questions.  \textbf{Balanced24} receives eight chronological frames
per call, so eight calls cover the 64 timestamps.  It writes one line per frame
under the following instruction:
\begin{quote}\small
Questions are hidden. Build a reusable factual record from these chronological
source frames. Output exactly one physical line for every supplied source
frame, in the same order, using this schema: Frame
\textless{}index\textgreater{} \textbar{} entities:
\textless{}visible people and objects\textgreater{} \textbar{} action/state:
\textless{}visible actions and state\textgreater{} \textbar{} visible text:
\textless{}legible text or none\textgreater{}. Do not merge, skip, repeat, add headings,
or use frame ranges. Record only visible facts and do not speculate. Each line
contains at most 24 words: at most five entities, eight words for action/state,
and eight words of visible text. Use ``none'' when a field has no visible
content.
\end{quote}
Writing uses temperature zero and a 1,024-token output ceiling per call.
Missing lines are requested again up to three times, accepted lines are never
shortened or rewritten, and every writer call counts toward construction cost.
The 64 accepted lines, each with its sampled time, are reused for every
question.  The \textbf{Detail24} ablation changes only the final priority
instruction: ``Within the same length limit, prioritize object contents,
counts, colors, spatial relations, package or label details, and exact visible
text. If space is tight, omit generic scene descriptions before these
details.''

All paths share the following answer prompt:
\begin{quote}\small\raggedright
Answer using only the supplied visual input or repository.\\
Return exactly one JSON object:\\
\{"answer":"A",\allowbreak
"evidence\_spans":[\{"role":"direct\_answer",\allowbreak
"start\_frame":0,\allowbreak
"end\_frame":0,\allowbreak
"key\_frames":[0],\allowbreak
"evidence":"at most 20 words"\}],\allowbreak
"rationale":"at most 20 words"\}\\
The value "A" above is only a format example. Replace it with exactly one uppercase\\
option label. Never copy a list such as "A|B|C" into the answer field.\\
Frame indices refer to the 64 supplied source frames and are zero-based. Cite only
evidence that supports the answer. If evidence is insufficient, return an empty
evidence\_spans list.
\end{quote}

\subsection{Answer Validity}
\label{sec:answer-format-exclusions}

A response is valid only if its JSON \texttt{answer} field contains exactly one
option label that agrees with the normalized prediction.  Empty, unparseable,
out-of-range, and multi-label answers such as \texttt{A|B} are invalid, and so
is a response whose evidence field cannot be parsed.  When either side of a
pair is invalid, we drop that question from both accuracy values but keep all
of its computation in the cost.  Valid-output rates are 94.4--100\% for Text,
91.1--100\% for Frames and Visual, and 90.7--100\% for Prefix State.  Counting
invalid answers as wrong instead changes the Text--Frames Flip from
19.7--28.7\% to 20.6--28.8\% and leaves its Loss at 10.0--15.7\%.  Under
the same rule, the Loss of Visual and Prefix State stays at or below
2.0\%.

\subsection{Global-Summary Text Ablation}
\label{sec:global-summary-method}

This ablation asks whether a video-level summary closes the gap between Text
and Frames.  A separate call rereads the same 64 frames with all questions
hidden and writes one line of at most 128 words.  \textbf{Added} prepends this
line to the Balanced24 records.  On CG MQ-50 we also test \textbf{Compact}, a
lossless and shorter serialization of the Balanced24 fields, and
\textbf{Fixed}, which adds the summary to Compact while staying no longer than
Balanced24 in words, UTF-8 bytes, and first-request input tokens.  Within the
ablation, timestamps, questions, checkpoints, answer prompt, and decoding
settings are fixed, and every summary call counts toward construction cost.
When four attempts still exceed the length limit, a deterministic fallback
shortens the summary.  The fallback was needed for 38 of 300 CG model--video
units and for 6 of 306 HD units, all of the latter from Gemma-3-12B-IT.
After excluding these units, the pooled CG Flip relative to Frames is 21.3\%
for Added and 22.0\% for Fixed.

Unlike Figure~\ref{fig:rq1-transitions}, the Flip and Loss rates in
Table~\ref{tab:global-summary} are computed relative to the Frames
condition rerun specifically for this ablation.  Added has a Flip of 18.6--29.6\% and a Loss of 9.7--17.0\%,
so the summary does not bring any model-dataset pair close to the 3\%
retention limit.

\begin{table}[!htbp]
\centering
\caption{\textbf{Text designs versus Frames in the summary ablation.}
Each cell reports Flip / Loss (\%) over valid pairs, computed relative to
the corresponding Frames run in the same ablation setting.
Fixed is evaluated only on CG MQ-50.}
\label{tab:global-summary}
\scriptsize
\setlength{\tabcolsep}{3pt}
\begin{tabular*}{\columnwidth}{@{\extracolsep{\fill}}llccc@{}}
\toprule
Data & Model & Balanced24 & Added & Fixed \\
\midrule
CG & Qwen3-VL-8B & 19.7/12.5 & 18.6/12.0 & 20.9/12.7 \\
   & InternVL3-8B & 23.5/14.0 & 22.6/14.0 & 24.1/13.8 \\
   & Gemma-3-12B-IT & 20.4/10.4 & 21.0/9.7 & 20.4/11.0 \\
   & Qwen3-VL-4B & 22.4/16.4 & 23.0/17.0 & 20.7/16.0 \\
   & InternVL3.5-8B & 24.2/15.7 & 23.0/14.7 & 26.7/17.8 \\
   & MiniCPM-V-4.5-8B & 22.7/11.7 & 20.8/11.1 & 21.7/12.3 \\
\midrule
HD & Qwen3-VL-8B & 21.6/11.9 & 21.8/11.3 & n/a \\
   & InternVL3-8B & 28.7/13.5 & 29.6/13.5 & n/a \\
   & Gemma-3-12B-IT & 24.5/10.0 & 24.2/10.3 & n/a \\
   & Qwen3-VL-4B & 24.3/12.6 & 24.5/12.1 & n/a \\
   & InternVL3.5-8B & 22.1/11.5 & 22.6/11.8 & n/a \\
   & MiniCPM-V-4.5-8B & 20.6/10.5 & 20.7/10.6 & n/a \\
\bottomrule
\end{tabular*}
\end{table}

\subsection{Models, Serving, and Accounting}

Table~\ref{tab:model-runtime-config} lists the checkpoints.  We run BF16
weights without quantization and let vLLM~\cite{kwon2023vllm} choose the
KV-cache data type.  Each
run uses one NVIDIA L40S GPU, 12 CPU cores, and 112 GiB of host memory, and
each server handles one request at a time.  Qwen uses its released image
processor, Gemma uses the released processor for its SigLIP vision
encoder~\cite{zhai2023siglip}, and InternVL uses one 448-by-448 patch per
timestamp.
All four paths of a model--dataset pair share one serving configuration, so we
compare token counts and retained sizes only within a pair.

\begin{table}[!t]
\centering
\caption{\textbf{Hugging Face repositories for the evaluated models.}}
\label{tab:model-runtime-config}
\scriptsize
\setlength{\tabcolsep}{1.0pt}
\begin{tabular*}{0.8\columnwidth}{@{\extracolsep{\fill}}ll@{}}
\toprule
Model & Hugging Face repository \\
\midrule
Qwen3-VL-8B & Qwen/Qwen3-VL-8B-Instruct \\
InternVL3-8B & OpenGVLab/InternVL3-8B \\
Gemma-3-12B-IT & google/gemma-3-12b-it \\
Qwen3-VL-4B & Qwen/Qwen3-VL-4B-Instruct \\
InternVL3.5-8B & OpenGVLab/InternVL3\_5-8B \\
MiniCPM-V-4.5-8B & openbmb/MiniCPM-V-4\_5 \\
\bottomrule
\end{tabular*}
\end{table}

Frames and Text disable vLLM prefix caching and the multimodal processor cache.
Visual keeps prefix caching off but caches the encoder outputs of the 64
frames, so later questions skip visual encoding but still run language-model
prefill.  Prefix State enables both caches, so later questions also reuse the
KV pages of the shared prefix.  For Visual and Prefix State, construction is a
question-blind one-token warm-up request through the same server, and its cost
includes visual encoding, prefill, and request overhead.  We charge GPU time for
every model call in construction and answering, including repairs, and exclude
model loading, data transfer, and queueing.  Retained size counts each backing
allocation once.  Text counts the UTF-8 answer repository, including each
frame's time field once; Visual counts the unique storage of the 64 cached
encoder outputs; and Prefix State adds the reached KV pages to Visual.  These
sizes exclude allocator overhead and unused server capacity.

\subsection{Frame Sampling and Selectors}

Uniform-$b$ selects $b$ frame indices evenly from the first to the last frame
and records their timestamps, and all paths in a comparison share this list.
On HD-EPIC, we follow the released evaluator, which turns \texttt{TIME} and
\texttt{BBOX} tags in a question into text; these tags never retrieve
additional frames.

\paragraph{Chunked MaxVol.}
The selector scans one frame per second and encodes each candidate with a
frozen CLIP ViT-L/14 at 336 pixels~\cite{radford2021clip}.  For a budget $K$,
it splits the candidates into $K/2$ contiguous blocks and picks two frames per
block with square MaxVol on the first two left singular vectors.  Unlike
published MaxInfo~\cite{li2026maxinfo}, which can return a variable number of
frames, our adaptation always returns exactly $K$ frames.  The scan runs once
per video, and its cost is charged once per video.

\paragraph{AKS.}
AKS~\cite{tang2025aks} rescans the same one-frame-per-second stream for every
question and scores frames with a frozen BLIP~\cite{li2022blip} image-text matching model, using
the released recursion settings (maximum depth 5, threshold 0.8, and candidate
ratio 1).  Its full scan and selection are charged to every question.  For both
selectors, each frame budget is charged as if it were run alone.

\subsection{Statistics and Repeated Runs}
\label{sec:run-to-run}

Confidence intervals use 10,000 video-level bootstrap samples.  A sampled video
keeps all of its questions and shared construction, and the 2.5th and 97.5th
percentiles form the 95\% interval.  These intervals describe variation across
videos, not across repeated runs.

Decoding uses temperature zero, so instead of sampling seeds we compare repeated
runs of the same path with the same answer prompt.  Across 13 such pairs, nine
on HD-EPIC and four on CG MQ-50, the run-to-run Flip is 0--2.8\%, and GPU time
per video changes by $-7.4\%$ to $+2.0\%$.  Other archived reruns
also changed the answer prompt or runtime settings, so we do not treat them as
repeated runs.  In 10 of the 36 within-model path-cost comparisons, the cost
gap is smaller than the largest batch difference observed for the paths
involved.  Recomputing the six HD-EPIC endpoints with only historical or only
current batches changes no endpoint choice.

Permuting the question order within each video 10,000 times for Qwen3-VL-8B and
Gemma-3-12B-IT on both datasets never changes which paths pass: Text passes in
none of the 40,000 orders, whereas Visual and Prefix State pass in all of them.
The permuted check uses the paired point estimate of the accuracy change rather
than a new bootstrap bound for every order.

\subsection{Scope and Limitations}

Hardware, frame or storage budgets, question mix, and the Text writer may
change the absolute cost rankings. The clue-window diagnostic cannot separate
writer omissions from information that the answer model failed to use.

%% file: sec/B_additional_experiments.tex
\section{Additional Experimental Results}
\label{app:additional-results}

\subsection{Answer Changes and Question Types}
\label{sec:text-flip-uncertainty}
\label{sec:question-types}

Across the 12 main model--dataset pairs, the Text--Frames Flip is
19.7--28.7\% of jointly valid questions, and the lower ends of the 95\%
video-bootstrap intervals range from 14.8\% to 26.2\%.

For each model--dataset pair, we test whether question type is associated with
the Frames-to-Text transition (loss, gain, or no change) using Pearson's
$\chi^2$ statistic and 20,000 within-video permutations, with one
Benjamini--Hochberg correction~\cite{benjamini1995fdr} across the 12 tests.  The
permutation keeps each video's types and transition totals, which controls for
video difficulty but leaves little power for rare types.
After the correction, the association holds for all six HD-EPIC models
($q_{\rm BH}\leq0.005$) and for none of the six CG-Bench models
($q_{\rm BH}\geq0.106$).

Question-type weights can also change the aggregate result.  For Gemma-3-12B-IT on
HD-EPIC, Text gains 4.6 points under the cohort's question mix and 5.5 points
under the official type frequencies.  When every official type receives equal
weight, however, the difference is $-0.1$ points, with a 95\% interval from
$-3.6$ to $+3.3$.

\begin{table*}[!t]
\centering
\caption{\textbf{All four paths at 32 and 64 frames on CG MQ-50.}
$G_6$ and Save are GPU-seconds per video through six questions, with Save
relative to Frames; Cross is the sustained compute crossing, and -- means no
crossing by $m=6$. Accuracy, Flip, and Loss are
percentages.}
\label{tab:uniform32}
\scriptsize
\setlength{\tabcolsep}{2.0pt}
\resizebox{\textwidth}{!}{%
\begin{tabular}{llrrrrrr@{\hspace{5pt}}rrrrrr}
\toprule
& & \multicolumn{6}{c}{32 frames} & \multicolumn{6}{c}{64 frames} \\
\cmidrule(lr){3-8}\cmidrule(lr){9-14}
Model & Path & $G_6$ & Save & Acc. & Flip & Loss & Cross & $G_6$ & Save & Acc. & Flip & Loss & Cross \\
\midrule
\multirow{4}{*}{Qwen3-VL-8B} & Frames & 40.7 & +0.0 & 36.3 & 0.0 & 0.0 & -- & 79.5 & +0.0 & 39.0 & 0.0 & 0.0 & -- \\
 & Text & 46.8 & -6.1 & 29.7 & 20.2 & 13.5 & -- & 79.3 & +0.2 & 34.0 & 19.7 & 12.5 & 6 \\
 & Visual & 35.8 & +4.9 & 36.3 & 0.0 & 0.0 & 3 & 69.5 & +10.0 & 39.0 & 0.0 & 0.0 & 3 \\
 & Prefix State & 22.8 & +17.9 & 35.3 & 1.0 & 1.0 & 2 & 32.8 & +46.7 & 40.0 & 1.4 & 0.3 & 1 \\
\midrule
\multirow{4}{*}{InternVL3-8B} & Frames & 35.4 & +0.0 & 36.3 & 0.0 & 0.0 & -- & 110.1 & +0.0 & 36.7 & 0.0 & 0.0 & -- \\
 & Text & 32.3 & +3.1 & 29.3 & 18.6 & 13.2 & 6 & 57.0 & +53.1 & 33.0 & 23.7 & 14.1 & 3 \\
 & Visual & 34.7 & +0.7 & 36.7 & 0.0 & 0.0 & 5 & 115.4 & -5.3 & 36.7 & 0.0 & 0.0 & -- \\
 & Prefix State & 30.8 & +4.6 & 37.3 & 1.7 & 0.3 & 3 & 106.8 & +3.3 & 37.0 & 0.3 & 0.0 & 5 \\
\midrule
\multirow{4}{*}{Gemma-3-12B-IT} & Frames & 37.0 & +0.0 & 25.7 & 0.0 & 0.0 & -- & 62.8 & +0.0 & 27.0 & 0.0 & 0.0 & -- \\
 & Text & 56.5 & -19.4 & 24.3 & 23.1 & 12.4 & -- & 95.9 & -33.1 & 26.3 & 20.4 & 10.4 & -- \\
 & Visual & 27.7 & +9.4 & 25.7 & 0.0 & 0.0 & 2 & 40.7 & +22.2 & 26.3 & 0.7 & 0.7 & 2 \\
 & Prefix State & 20.3 & +16.7 & 26.0 & 0.3 & 0.0 & 2 & 27.2 & +35.7 & 26.7 & 1.0 & 0.7 & 1 \\
\bottomrule
\end{tabular}}
\end{table*}

\subsection{Clue-Window Diagnostic}

For each CG-Bench question, an oracle sampler places 64 positions inside the
question's annotated clue intervals.  Frames and Text receive the same
positions, and the writer remains question-blind.
Table~\ref{tab:evidence-window-full} compares the Text-minus-Frames gap under
uniform and clue-window sampling on the questions that are valid in all four
cells.  Clue-window sampling raises the accuracy of both paths, but all three
interaction intervals include zero, so the diagnostic does not show a
consistent narrowing of the gap.  Because the oracle uses the current
question's annotation, we exclude it from all cost comparisons.

\begin{table}[!htbp]
\centering
\caption{\textbf{Text-minus-Frames gaps under uniform and clue-window sampling.}
Values are percentage points with 95\% video-bootstrap intervals. A positive
interaction means clue-window sampling narrows the gap.}
\label{tab:evidence-window-full}
\scriptsize
\setlength{\tabcolsep}{2.5pt}
\resizebox{\columnwidth}{!}{%
\begin{tabular}{llll}
\toprule
Model & Uniform64 & Clue window64 & Interaction \\
\midrule
Qwen3-VL-8B & $-5.1\,[-9.8,-0.7]$ & $-11.6\,[-16.5,-6.8]$ & $-6.5\,[-13.2,+0.3]$ \\
InternVL3-8B & $-4.5\,[-10.1,+1.1]$ & $-2.8\,[-9.2,+3.5]$ & $+1.7\,[-5.8,+9.1]$ \\
Gemma-3-12B-IT & $-0.3\,[-6.4,+5.7]$ & $-3.0\,[-9.7,+3.7]$ & $-2.7\,[-10.1,+5.1]$ \\
\bottomrule
\end{tabular}}
\end{table}

\begin{table*}[!t]
\centering
\caption{\textbf{MaxVol and AKS across backbones at 64 frames.}
Accuracy uses questions with valid answers from both selectors. The interval
is a 95\% video-bootstrap interval for AKS minus MaxVol. All $\Delta$ instead
counts invalid answers as wrong. The GPU-time ratios include selection and
answering over all evaluated questions: six per CG-Bench video and all 1,871
questions across the 51-video HD-EPIC-51 cohort. For its GPU-time totals,
Table~\ref{tab:selector-sweep} instead uses the same six-question CG-Bench
horizon but only the first 20 HD-EPIC questions per video.}
\label{tab:selector-transfer-k64}
\scriptsize
\setlength{\tabcolsep}{3.5pt}
\begin{tabular*}{\textwidth}{@{\extracolsep{\fill}}llrrrrrrr@{}}
\toprule
& & & \multicolumn{2}{c}{Accuracy (\%)} & & & & \\
\cmidrule(lr){4-5}
Data & Model & Valid/All & MaxVol & AKS & $\Delta$ & 95\% CI & All $\Delta$ &
\makecell{GPU ratio\\AKS/MaxVol} \\
\midrule
\input{figures/appendix/selector_rows.tex}\end{tabular*}
\end{table*}

\subsection{Writing-Priority Ablation}

Table~\ref{tab:prompt-ablation-full} shows that the Flip between Detail24 and
Balanced24 is 6.8--9.8\% of valid pairs,
although the realized length changes by only 0.4--2.8\% and all six intervals
for the accuracy change include zero.

\begin{table}[!htbp]
\centering
\caption{\textbf{Balanced24 versus Detail24 at the same requested length.}
Pairs are valid under both writers, $\Delta$Acc is Detail24 minus Balanced24,
and Flip is the fraction with different correctness, each with a 95\%
video-bootstrap interval. Valid-output rates are 98.7--100\% for both writers.}
\label{tab:prompt-ablation-full}
\scriptsize
\setlength{\tabcolsep}{1.6pt}
\resizebox{\columnwidth}{!}{%
\begin{tabular}{llrrrr}
\toprule
Data & Model & Words/frame & Pairs & $\Delta$Acc & Flip (\%) \\
\midrule
CG & Qwen3-VL-8B & $19.28{\to}19.49$ & 296/300 & $+1.4$ [$-1.0$, $+3.7$] & 6.8 [4.0, 9.8] \\
   & InternVL3-8B & $17.49{\to}17.66$ & 299/300 & $+1.3$ [$-1.7$, $+4.3$] & 8.7 [5.7, 12.0] \\
   & Gemma-3-12B-IT & $16.34{\to}16.79$ & 297/300 & $-1.7$ [$-5.7$, $+2.7$] & 9.8 [6.4, 13.4] \\
\midrule
HD & Qwen3-VL-8B & $19.32{\to}19.40$ & 1,853/1,871 & $+0.0$ [$-1.2$, $+1.2$] & 8.6 [7.4, 9.9] \\
   & InternVL3-8B & $17.41{\to}17.74$ & 1,869/1,871 & $-0.5$ [$-2.0$, $+1.0$] & 9.5 [8.0, 11.1] \\
   & Gemma-3-12B-IT & $17.09{\to}17.23$ & 1,854/1,871 & $-1.5$ [$-3.0$, $+0.1$] & 8.9 [7.5, 10.4] \\
\bottomrule
\end{tabular}}
\end{table}

\subsection{Cumulative GPU Time for All Six Backbones}
\label{sec:new-model-rq2}

\begin{figure}[!htbp]
\centering
\includegraphics[width=\linewidth]{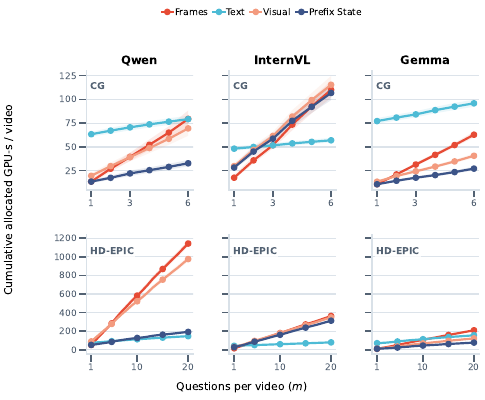}
\caption{\textbf{Cumulative GPU time for Qwen3-VL-8B, InternVL3-8B, and
Gemma-3-12B-IT.}
Rows are datasets; bands are 95\% video-bootstrap intervals.}
\label{fig:cumulative-cost-full}
\end{figure}

\begin{figure}[!htbp]
\centering
\includegraphics[width=\linewidth]{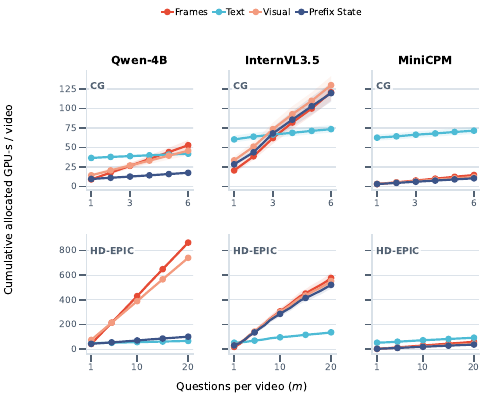}
\caption{\textbf{Cumulative GPU time for Qwen3-VL-4B, InternVL3.5-8B, and
MiniCPM-V-4.5-8B.}
Rows are datasets; bands are 95\% video-bootstrap intervals.}
\label{fig:new-model-cumulative-cost}
\end{figure}

Figures~\ref{fig:cumulative-cost-full} and~\ref{fig:new-model-cumulative-cost}
show the cumulative GPU time of all four paths for the six backbones, and
Table~\ref{tab:selection-paths} in the main paper gives the resulting choices.
If Visual is charged only for the incremental work of its warm-up, measured as
a cold request minus a cache-hit replay, its crossing moves earlier in five of
the six model--dataset pairs of Figure~\ref{fig:cumulative-cost-full}, but no
endpoint choice changes.

\begin{table*}[!t]
\centering
\caption{\textbf{HER-MQ10 results for two backbones.}
Valid counts each path's outputs; Pair counts questions valid for Frames and
the listed path. $\Delta$Acc, Loss, and Flip use those pairs. $B$ and $G_{10}$ are GPU-seconds per video, Save
is relative to Frames, and the final columns give the sustained crossings.}
\label{tab:herbench-full}
\scriptsize
\setlength{\tabcolsep}{2.2pt}
\resizebox{\textwidth}{!}{%
\begin{tabular}{llrrrrrrrrrr}
\toprule
Model & Rep. & Valid & Pair & $\Delta$ [95\% CI] & Loss & Flip & $B$ & $G_{10}$ & Save & GPU from & All from \\
\midrule
Qwen3-VL-8B & Frames & 493/500 & 493/500 & $+0.00$ [$+0.00$, $+0.00$] & 0.0\% & 0.0\% & 0.0 & 479.4 & 0.0\% & -- & -- \\
Qwen3-VL-8B & Text & 494/500 & 487/500 & $-9.24$ [$-14.49$, $-4.29$] & 16.2\% & 23.2\% & 70.0 & 93.6 & 80.5\% & 2 & -- \\
Qwen3-VL-8B & Visual & 492/500 & 492/500 & $+0.00$ [$+0.00$, $+0.00$] & 0.0\% & 0.0\% & 42.4 & 421.2 & 12.1\% & 5 & 5 \\
Qwen3-VL-8B & Prefix State & 492/500 & 492/500 & $-0.20$ [$-0.61$, $+0.00$] & 0.2\% & 0.2\% & 42.2 & 104.1 & 78.3\% & 2 & 3 \\
InternVL3-8B & Frames & 499/500 & 499/500 & $+0.00$ [$+0.00$, $+0.00$] & 0.0\% & 0.0\% & 0.0 & 173.3 & 0.0\% & -- & -- \\
InternVL3-8B & Text & 493/500 & 492/500 & $-3.46$ [$-7.54$, $+0.61$] & 13.8\% & 24.2\% & 40.9 & 59.7 & 65.5\% & 3 & -- \\
InternVL3-8B & Visual & 498/500 & 498/500 & $+0.00$ [$+0.00$, $+0.00$] & 0.0\% & 0.0\% & 13.6 & 172.4 & 0.5\% & 10 & 10 \\
InternVL3-8B & Prefix State & 500/500 & 499/500 & $-0.80$ [$-2.00$, $+0.20$] & 1.2\% & 1.6\% & 13.6 & 155.4 & 10.3\% & 5 & 7 \\
\bottomrule
\end{tabular}}
\end{table*}

\subsection{Sensitivity to Both Quality Limits}
\label{sec:quality-threshold-sensitivity}

We vary the mean-accuracy limit $\epsilon$ over 0, 0.02, 0.03, and 0.05 and
the Loss limit $\delta$ over 0.02, 0.03, and 0.05.  Across this grid,
Text passes in none of the  12 model--dataset pairs, Visual in 6 to 10, and Prefix State in
1 to 12.  Changing $\delta$ alone changes no count, whereas $\epsilon$ decides
whether several Visual and Prefix State paths pass.

\subsection{Retained Size and Storage Caps}

\begin{figure}[!htbp]
\centering
\includegraphics[width=\linewidth]{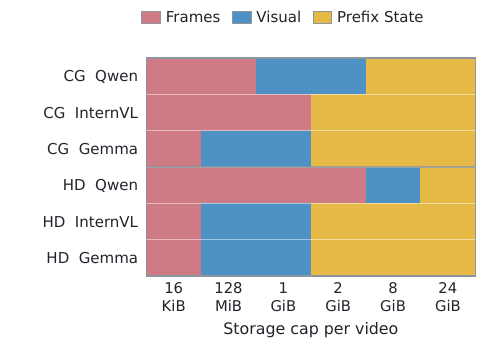}
\caption{\textbf{The best answer-preserving path depends on the storage budget.}
At each cap, color gives the lowest-GPU path among Frames and the reusable
paths that fit the per-video cap and pass both quality checks. Rows are the
three profiled models on CG-Bench (CG) and HD-EPIC (HD). Figure labels Qwen,
InternVL, and Gemma denote Qwen3-VL-8B, InternVL3-8B, and Gemma-3-12B-IT.}
\label{fig:storage-budget-selection}
\end{figure}

Table~\ref{tab:representation-size} gives the median retained payload per video
for the three backbones whose payloads we measured.
Figure~\ref{fig:storage-budget-selection} applies a per-video storage cap
before the GPU and quality checks.  Under the smallest cap only Frames passes,
because Text fails the retention check.  Visual is chosen in three or four of
the six model--dataset pairs at caps from 128~MiB to 1~GiB, and Prefix State is chosen in
four model--dataset pairs at 2~GiB and in all six at 24~GiB.  Storing the KV pages in 8
bits, with the Visual part unchanged, would reduce the Prefix State totals by
an estimated 41--46\%; we did not test whether this changes answers.

\begin{table}[!htbp]
\centering
\caption{\textbf{Median retained payload per video for the same 64 frames.}
Text counts serialized bytes including each timestamp once, Visual counts
unique encoder-output storage, KV only counts reached pages, and Prefix State
adds Visual and KV storage.}
\label{tab:representation-size}
\scriptsize
\resizebox{\columnwidth}{!}{%
\input{figures/appendix/payload_table.tex}}
\end{table}

\subsection{Published-Tariff Counterfactual}

We do not run GLM inference.  Instead, we apply GLM-5V-Turbo's published tariff
to the token counts recorded in our runs, including construction and the first
$m$ questions.  Below 32K input tokens, cache-miss input, cache-hit input, and
output tokens cost 5, 1.2, and 22 RMB per million tokens; longer requests cost
7, 1.8, and 26 RMB per million tokens~\cite{zhipu2026pricing,zhipu2026cache}.
``No cache'' prices every input token as a miss, whereas ``Observed hits''
discounts only the tokens served from our local prefix cache.  The calculation
does not reproduce GLM tokenization, image billing, or automatic cache
eligibility.  Under the published rates, Text is the cheapest endpoint path in
all six model--dataset pairs, and Table~\ref{tab:glm-tariff-crossing} gives the
sustained monetary crossings.

\begin{table}[!htbp]
\centering
\caption{\textbf{Sustained monetary crossing under the GLM tariff.}
Costs include construction and answering; $>H$ means no crossing by the
endpoint.}
\label{tab:glm-tariff-crossing}
\scriptsize
\setlength{\tabcolsep}{2.0pt}
\begin{tabular*}{\columnwidth}{@{\extracolsep{\fill}}lccc@{\hspace{4pt}}ccc@{}}
\toprule
& \multicolumn{3}{c}{No cache} & \multicolumn{3}{c}{Observed hits} \\
Model & Text & Visual & \makecell{Prefix\\State} & Text & Visual & \makecell{Prefix\\State} \\
\midrule
\multicolumn{7}{l}{\emph{CG, endpoint $H=6$}} \\
Qwen3-VL-8B & 2 & $>6$ & $>6$ & 2 & $>6$ & 2 \\
InternVL3-8B & 2 & $>6$ & $>6$ & 2 & $>6$ & 2 \\
Gemma-3-12B-IT & 2 & $>6$ & $>6$ & 2 & $>6$ & 2 \\
\midrule
\multicolumn{7}{l}{\emph{HD-EPIC, endpoint $H=20$}} \\
Qwen3-VL-8B & 1 & $>20$ & $>20$ & 1 & $>20$ & 2 \\
InternVL3-8B & 2 & $>20$ & $>20$ & 2 & $>20$ & 2 \\
Gemma-3-12B-IT & 2 & $>20$ & $>20$ & 2 & $>20$ & 2 \\
\bottomrule
\end{tabular*}
\end{table}

\subsection{All Four Paths at 32 and 64 Frames}

Table~\ref{tab:uniform32} repeats the four-path comparison with 32 uniform
frames on CG MQ-50 for Qwen3-VL-8B, InternVL3-8B, and Gemma-3-12B-IT.  Within
each frame budget, the paths share the videos, questions, timestamps,
checkpoints, answer prompt, and accounting rule.  At 32 frames, Text still has
a Flip of 18.6--23.1\% relative to Frames, whereas Visual and Prefix State
have a Flip of at most 1.7\%.  Storage was measured only at 64 frames.

\subsection{Representation Check with Chunked MaxVol}
\label{sec:maxvol-representation-check}

Under Chunked MaxVol-64 on CG MQ-50 with Qwen3-VL-8B, InternVL3-8B, and
Gemma-3-12B-IT, Text has a Flip of 17.1--22.4\% relative to Frames,
compared with at most 0.7\% for Visual or Prefix State.  Adding Text
to a retained state changes accuracy by $-1.40$ to $+2.46$ points relative to
the better component.  The resulting combination has a Loss of 3.19--5.24\%
relative to Frames, above the 3\% limit in all six combinations.

\input{sec/B_selector_transfer_k64}

\subsection{HER-MQ10}

Table~\ref{tab:herbench-full} reports the HER-MQ10 check.  All paths answer the
same 500 questions with one timestamp list per video, and 487--499 questions
are valid for both paths of each pair.

%% file: figures/appendix/selector_rows.tex
CG & Qwen3-VL-8B & 289/300 & 41.9 & 48.1 & $+6.2$ & $[+0.0,\,+12.5]$ & $+6.3$ & $2.33\times$ \\
CG & InternVL3-8B & 294/300 & 42.2 & 44.9 & $+2.7$ & $[-2.0,\,+7.1]$ & $+2.7$ & $2.09\times$ \\
CG & Gemma-3-12B-IT & 298/300 & 27.2 & 28.9 & $+1.7$ & $[-2.7,\,+6.0]$ & $+1.7$ & $2.72\times$ \\
CG & Qwen3-VL-4B & 299/300 & 42.5 & 46.8 & $+4.3$ & $[-1.3,\,+10.0]$ & $+4.3$ & $2.74\times$ \\
CG & InternVL3.5-8B & 300/300 & 38.7 & 45.7 & $+7.0$ & $[+1.3,\,+12.7]$ & $+7.0$ & $2.00\times$ \\
CG & MiniCPM-V-4.5-8B & 300/300 & 39.0 & 45.3 & $+6.3$ & $[+1.3,\,+11.7]$ & $+6.3$ & $4.56\times$ \\
\midrule
HD & Qwen3-VL-8B & 1835/1871 & 31.3 & 30.2 & $-1.1$ & $[-2.8,\,+0.6]$ & $-1.1$ & $4.40\times$ \\
HD & InternVL3-8B & 1847/1871 & 28.3 & 28.0 & $-0.2$ & $[-2.2,\,+1.8]$ & $-0.2$ & $3.44\times$ \\
HD & Gemma-3-12B-IT & 1870/1871 & 24.1 & 25.8 & $+1.7$ & $[-0.3,\,+3.8]$ & $+1.7$ & $5.24\times$ \\
HD & Qwen3-VL-4B & 1871/1871 & 26.2 & 27.7 & $+1.5$ & $[-0.2,\,+3.1]$ & $+1.5$ & $6.13\times$ \\
HD & InternVL3.5-8B & 1871/1871 & 28.5 & 28.3 & $-0.3$ & $[-2.3,\,+1.7]$ & $-0.3$ & $2.94\times$ \\
HD & MiniCPM-V-4.5-8B & 1871/1871 & 23.6 & 24.6 & $+1.0$ & $[-0.8,\,+3.0]$ & $+1.0$ & $13.66\times$ \\
\bottomrule

%% file: figures/appendix/payload_table.tex
\begin{tabular}{llrrrr}
\toprule
Dataset & Model & Text & Visual & KV only & Prefix State \\
\midrule
CG & Qwen & 9.7 KiB & 810.0 MiB & 3.74 GiB & 4.53 GiB \\
CG & InternVL & 8.7 KiB & 112.9 MiB & 973.0 MiB & 1.06 GiB \\
CG & Gemma & 7.7 KiB & 121.9 MiB & 1.40 GiB & 1.52 GiB \\
\midrule
HD & Qwen & 8.5 KiB & 3.78 GiB & 17.19 GiB & 20.97 GiB \\
HD & InternVL & 7.8 KiB & 112.9 MiB & 972.1 MiB & 1.06 GiB \\
HD & Gemma & 7.2 KiB & 121.9 MiB & 1.40 GiB & 1.52 GiB \\
\bottomrule
\end{tabular}

%% file: sec/B_selector_transfer_k64.tex
\subsection{Frame Selectors across Backbones}
\label{sec:selector-transfer-k64}

The main paper sweeps 32, 64, and 128 frames with Qwen3-VL-8B.  Here we repeat
the matched 64-frame MaxVol--AKS comparison across all six backbones.  Both
selectors use the Frames answer path, and AKS is charged for rerunning its
question-aware search for every question.  Because InternVL3.5 sometimes
exhausted the 384-token limit while writing the evidence field, its runs allow
one schema-constrained repair request with the same frames and options, and
both calls are charged.  Table~\ref{tab:selector-transfer-k64} reports both the
paired-valid comparison and a check that counts invalid answers as wrong.
\input{figures/appendix/selector_summary.tex}

Table~\ref{tab:uniform-aks-reference} compares AKS with uniform sampling for
Qwen3-VL-8B on CG-Bench, counting invalid answers as wrong.

\begin{table}[!htbp]
\centering
\caption{\textbf{Uniform reference for AKS on CG-Bench.}
All questions are counted, with invalid answers treated as wrong.
The Uniform values come from the dedicated selector-sweep runs and are
therefore separate from the four-path representation runs in
Table~\ref{tab:uniform32}.}
\label{tab:uniform-aks-reference}
\scriptsize
\setlength{\tabcolsep}{5pt}
\begin{tabular}{rrrr}
\toprule
Frames & Uniform & AKS & $\Delta$ \\
\midrule
32 & 36.0 & 45.7 & $+9.7$ \\
64 & 41.0 & 48.3 & $+7.3$ \\
128 & 47.0 & 54.7 & $+7.7$ \\
\bottomrule
\end{tabular}
\end{table}

%% file: figures/appendix/selector_summary.tex
At 64 frames, AKS has the higher paired-accuracy point estimate for all six
backbones on CG-Bench and three of six on HD-EPIC. The 95\% interval excludes
zero for two CG-Bench backbones and no HD-EPIC backbone. Including selection
and answering, AKS uses $2.00$--$13.66\times$ as much total GPU time as MaxVol
across the 12 model--dataset pairs (Table~\ref{tab:selector-transfer-k64}).